\documentclass[runningheads]{llncs}
\usepackage[T1]{fontenc}
\usepackage{commands}
\usepackage{tikz}
\usetikzlibrary{positioning}
\usepackage{amsmath}
\usepackage{amsfonts}
\usepackage{amssymb}
\usepackage{graphicx}
\begin{document}
\title{Logic-Based Ethical Planning}
\author{Umberto Grandi\inst{1}\orcidID{0000-0002-1908-5142} \and
Emiliano Lorini\inst{1}\orcidID{0000-0002-7014-6756} \and
Timothy Parker\inst{1}\orcidID{0000-0002-5594-9569} \and
Rachid Alami\inst{2}\orcidID{0000-0002-9558-8163}}
\authorrunning{U. Grandi et al.}
\institute{IRIT, CNRS and University of Toulouse, France \and
LAAS-CNRS, France}
\maketitle              % typeset the header of the contribution
\begin{abstract}
In this paper we propose a framework for ethical decision making in the context of planning, with intended application to robotics. 
We put forward a compact but highly expressive language for ethical planning that combines linear temporal logic with lexicographic preference modelling.
This original combination allows us to assess plans both with respect to an agent's values and their desires, introducing the novel concept of the morality level of an agent and moving towards multi-goal, multi-value planning. 
We initiate the study of computational complexity of planning tasks in our setting, and we
discuss potential applications to robotics.

\keywords{KR and ethics \and Linear temporal logic \and Compact preference representation \and Robotics}
\end{abstract}
\section{Introduction}
In 
ethical planning
the planning agent has to find a plan for
promoting a certain number of ethical values.
The latter include both abstract values such as justice, fairness, reciprocity, equity, respect for human integrity and more concrete ones such as ``greenhouse gas emissions are reduced''.
Unlike classical planning  in which
the goal to be achieved is unique, in ethical
planning the agent can have multiple and possibly conflicting values,
that is, values that 
cannot be concomitantly satisfied.
It is typical
of ethical planning
the problem of facing a moral struggle
which is ``...provoked by inconsistencies between value commitments and information concerning the kinds of decision problems which arise...'' \cite[p. 8]{LeviBook1990}.
Consequently,
in ethical planning the agent needs to evaluate and compare the ideality (or goodness) of different plans depending on
how many and which  values are promoted by each of them. 

%\elnote{Added previous paragraph clarifying the concept of ethical planning. To be harmonized with the rest of the paper; }
%
%[ONGOING WORK, STILL SKETCHY - UG]

In this paper our intended application field is that of robotics. Including ethical considerations in robotics planning requires (at least) three steps.
First, identify ethically sensitive situations in the robotics realm, and how are these situations represented. 
%A classical model here is planning, in which the world is described by variables and a sequence of actions needs to be selected to achieve a certain goal. 
Planning seems to be the first candidate in which to include ethical considerations, thus we assume that values or ethical judgments are expressed about the results of plans.
Second, design a language to express such values, taking in mind that they can be, and often are, potentially conflicting in multiple ways: among values, between a value and a goal, or between a value and good practices.
Such a value representation language needs to be compact and computationally tractable.
Third, complete the picture of ethical planning by designing algorithms that compare plans based on the ethical values. 
%Computational complexity is key here to ensure applicability of the approach.

%A number of work in the recent literature on ethics in AI touched some of these aspects, but to the best of our knowledge none of them proposed a full-fledged model of ethical decision-making.
%Much work focused on how ethical vales can be written in a language, such as CP nets or similar CITE. 
%Complementary to that, some work studied how values can be learned or taught from the bottom up to a software agent. 
%Our intended application is robotics, thus our examples and definitions are tailored to it (hence the second approach might be less applicable).

In this paper we put forward a framework for ethical planning based on a simple temporal logic language to express both an agent's values and goals. 
For ease of exposition we focus on single-agent planning with deterministic sequential actions in a known environment.
Our model borrows from the existing literature on planning and combines it in an original way with research in compact representation languages for preferences. 
The latter is a widely studied topic in knowledge representation, where logical and graphical languages are proposed to represent compactly the preferences of an agent over a combinatorial space of alternatives, often described by means of variables. 
In particular, we commit to a prioritised or lexicographic approach
%in dealing with multiple values and goals, 
to solve the possible arising inconsistencies among include goals, desires, and good practices in a unified planning model.

\section{Related Work}

There is considerable research in the field of ethics and AI, see \citet{sep-ethics-ai} for a general overview. Popular ethical theories for application are consequentialism, 
%(good actions are those that have good overall consequences) \cite{Jenkin},
deontology, 
%(good actions are those that follow some pre-existing set of rules) \cite{Powers} 
and virtue ethics.\footnote{ 
%(good actions are those that are in line with moral virtues) \cite{Vallor}.
See \citet{HandbookEthics2007} for a philosophical introduction, and \citet{Jenkin}, \citet{Powers}, and \citet{Vallor} for a discussion of these three theories in robotics.}
Our approach should be able to work with any notion of ``good actions'' but is probably a most natural fit for %so-called
pluralistic consequentialism
\cite{SenEthicsBook}.

While there is a lot of work at the theoretical/abstract level, there is comparatively less that examines how ethical reasoning in artificial agents could actually be done in practice. 
There are approaches both in terms of formal models \cite{Dennis} and allowing agents to learn ethical values \cite{GenEth}. 
%Furthermore, to our knowledge there is currently no work on combining ethical reasoning with planning problems. \noteug{Not true, see papers sent by Rachid.}
\citet{YuEtAlIJCAI2018} provides a recent survey of this research area.
The closest approaches to ours are
the recent work on $(i)$ 
logics for ethical reasoning and $(ii)$ the combination
of a compact representation language, such as conditional preference networks, with decision-making in an ethically sensitive domain.
The former are based on different methodologies including 
event calculus (ASP) \cite{DBLP:conf/atal/BerrebyBG17},
epistemic logic
and preference logic
\cite{LoriniMoral,DBLP:conf/atal/Lorini21},
BDI (belief, desire, intention) agent language \cite{DennisFSW16}, 
classical higher-order logic (HOL) \cite{DBLP:journals/ai/BenzmullerPT20}.
The latter was presented in ``blue sky'' papers \cite{LoreggiaEtAl2017,RossiMatteiAAAI2019} complemented with a technical study of distances between CP-nets \cite{LoreggiaEtAlAAMAS2018} and, more recently, with an empirical study on human ethical decision-making \cite{AwadEtAlarXiv2022}.
CP-nets are a compact formalism to order states of the world described by variables. 

We take inspiration from these lines of work, but depart from them under two aspects.
First, robotics applications are dynamical ones, and ethical principles must be expressed over time. Hence, unlike existing logics
for ethical reasoning, our focus is on a specification language
for values based on linear temporal  logic.
%$\ltllogic$. 
Second, ethical  
decision-making in robotic applications
requires mixing potentially conflicting values with 
desires of the agent and to express the notion of plan, and CP-nets alone are not sufficient.

In the field of robotics, there are approaches to enabling artificial agents to compute ethical plans.
The evaluative component, which consists in assessing the  ``goodness'' of an action
or a plan in relation to the robot's values,  
is made explicit by 
\citet{DBLP:journals/pieee/ArkinUW12} and \citet{DBLP:journals/cogsr/VanderelstW18}. \citet{EvansK} focuses on a collision scenario involving an autonomous vehicle, proposing to prioritise the ethical claims depending on the situation, e.g. by giving more priorities to the claims of the more endangered agents.
%where each individual has a ``claim'' on the vehicle's  behaviour (amongst other things, those in more danger have stronger claims) and instructs the vehicle to act in accordance with the strongest claim. This is somewhat similar to our lexicographic ordering of sets of values, but this is a model optimised for a specific type of ethical dilemma, whereas ours is much more general, particularly given the expressivity of $\ltllogic$. 
Related work explores the design of planning algorithms designed to help robots produce socially acceptable plans by assigning weights to social rules \cite{AliliEtAl2008} . 
%It does this by assigning penalty weights to plans that violate various social rules, and attempts to generate the plan with the lowest weight that still achieves the goal. While this system is designed to handle social rules, in principle this architecture could handle at least some ethical rules.
%Evaluating values using weights can lead to quite different behaviour, as it means that sufficiently many lower-priority values can ``override'' a higher-priority value. Whether or not this is desirable will depend a lot on the specific values and scenario being considered. 
%[TODO: look for the publications at the workshops at AAAI on machine ethics to see if anything similar has been published. See also the survey paper at IJCAI which is a good starting point \cite{YuEtAlIJCAI2018}]

\section{Model  }

In this section,
we present the formal model 
of ethical evaluation and planning
which consist, respectively, in
comparing the goodness of plans
and in finding the best plan relative to a given base of ethical values.

\subsection{LTL Language   }

Let $\propset$
be a countable
set of atomic propositions
and let 
$\actset$
be a finite non-empty set of action names.
Elements
of $\propset$
are noted $p, q, \ldots$,
while elements of 
$\actset$
are noted $a, b , \ldots$. We assume the existence of a special action $\nullact$.
The set of states  is $\stateset = 2^{\propset }$
with elements $s,s', \ldots$

In order to represent the agent's
values,
we introduce
the language of $\ltllogic$
(Linear Temporal Logic over Finite Traces) \cite{Pnueli1977,GiacomoV13}, noted 
$\langlogic_{\ltllogic}(\propset )$ 
(or 
%simply 
$\langlogic_{\ltllogic}$), defined by the following grammar:
\begin{center}\begin{tabular}{lcl}
  $\phi $  & $\bnf$ & $ p   \mid  \neg\phi \mid \phi_1  \wedge \phi_2   \mid \nexttime \phi \mid
\until { \phi_1    } { \phi_2    },  $\\
\end{tabular}\end{center}
with 
$p$ ranging  over $\propset$.
$\nexttime$
and $\until {     } {     }$
are the  operators
``next''
and ``until'' of $\ltllogic$. 
Operators
``henceforth'' ($\henceforth$)
and ``eventually'' ($\eventually$)
are defined in the usual way:
$\henceforth \phi \defin \neg ( \until {   \top   } {  \phi     }) $
and 
$\eventually \phi  \defin \neg  \henceforth \neg  \phi $. 
The propositional logic
fragment of
$\langlogic_{\ltllogic}$ 
is noted
$\langlogic_{\proplogic}$
and is defined in the usual way.
%as follows:
%\begin{center}\begin{tabular}{lcl}
%$\phi    $  & $\bnf$ & $ p \mid 
%\neg\phi    \mid \phi_1  \wedge \phi_2  .  $
%\end{tabular}\end{center}
We will use 
$\langlogic_{\proplogic}$
to describe 
the  effect
preconditions of the agent's actions. 

 \subsection{Histories  }

The notion of history
is needed for interpreting  formulas in $\langlogic_{\ltllogic}$.
We define a $k$-history 
to be  a pair $\history= (\historyst, \historyact)$
with 
\begin{align*}
   \historyst:
    [0,k] \longrightarrow \stateset \text{ and } \historyact: [1,k] \longrightarrow \actset.
\end{align*}
A history specifies 
the actual configuration of
the environment  at
a certain  time point and the action executed by the agent
that leads  to the next state. The set of $k$-histories is noted $\historyset{k}$.
The set of histories is $\historyset{}=\bigcup_{k \in \nat }\historyset{k}$.  
Semantic
interpretation of
formulas in
$\langlogic_{\ltllogic}$
relative to a $k$-history 
$\history \in \historyset{}$ and a time point
$t \in [0,k] $ goes as follows (we omit
boolean cases which are defined as usual):
\begin{alignat*}{2}
  \history,  t &\models  p  & ~\IFF~ & 
  p \in \historyst(t) , \\
 %    \history,  k &\models  \neg \phi & ~\IFF~ &  \history,  k \not \models   \phi ,\\
  % \history,  k &\models   \phi_1 \wedge\phi_2 & ~\IFF~ &  \history,  k \models   \phi_1  \AND
   %  \history,  k \models   \phi_2 ,\\
  \history,  t &\models \nexttime \phi & ~\IFF~ & t < k \AND \history,  t+1 \models   \phi, \\
   \history,  t &\models \until { \phi_1    } { \phi_2    }   & ~\IFF~ &
    \begin{aligned}[t]
      &\exists t' \geq t :   t' \leq k \AND \history,  t' \models \phi_2 \AND
    \\ %&&&
      &\forall t''   \geq t  :   \IF 
      t'' < t' ~\THEN~ \history,  t'' \models \phi_1.
    \end{aligned} 
\end{alignat*}

 \subsection{Action Theory }
We suppose actions in 
$\actset$ are described by an action theory 
$\effprecond=(\effprecondplus,\effprecondminus)$,
where 
$\effprecondplus$
and $\effprecondminus$
are, respectively, 
the  positive and  negative effect precondition
function:
\begin{align*}
   & \effprecondplus:
    \actset \times \propset
   \longrightarrow \langlogic_{\proplogic},\\
      & \effprecondminus :
    \actset \times \propset
   \longrightarrow \langlogic_{\proplogic}.
\end{align*}
The fact 
$\effprecondplus( a,p )$ 
guarantees 
that proposition $p$
will be \emph{true} in the next state when action $a$
is executed, while 
$\effprecondminus( a,p )$ 
guarantees 
that proposition $p$
will be \emph{false} in the next state  when action $a$
is executed.
We stipulate that
if $\effprecondplus( a,p )$ 
and
$\effprecondminus( a,p )$ 
are concomitantly true 
at a given state
and action $a$
is executed, 
then the truth value of $p$
will not change in the next state. The latter captures
an  inertial principle for fluents.
%The special symbol $\nullact$ in $\actset $ with $\effprecondplus ( \nullact,p )=\effprecondminus( \nullact,p )= \bot$ is of particular interest. It denotes the agent's inaction that keeps the environment unmodified. 

\begin{definition}[Action-compatible histories]
Let
$\effprecond=(\effprecondplus,\effprecondminus)$
be an action theory
and let $\history= (\historyst, \historyact)$ be a $k$-history.
We say $\history$
is compatible with $\effprecond$
if the following condition holds,
for every  $t \in [1,k]$
and for every $a \in  \actset $ :
\begin{align*}
~\IF~ \historyact(t)=&~a ~\THEN~    \\
 \historyst(t) =&
\Big( \historyst (t-1)  \setminus 
\big\{p  \in \propset \suchthat \history , t-1 \models 
\neg  \effprecondplus( a,p )  \wedge \\
& \effprecondminus( a,p ) \big\}\Big) \cup 
   \big\{p  \in \propset \suchthat \history  , t-1 \models 
  \effprecondplus( a,p )  \wedge \\
& \neg \effprecondminus( a,p ) \big\}. 
\end{align*}
The set of $\effprecond$-compatible histories
is noted $\historyset{}(\effprecond )$.
\end{definition}

 \subsection{Plans }

Let us now move from the notion
of action to the notion of plan.
Given $k \in \nat$, a $k$-plan
is a function 
\begin{align*}
    \plan: \{0, \ldots, k\} \longrightarrow \actset.
\end{align*}
The set of $k$-plans is noted $\planset{k}$.
The set of plans is $\planset{}=\bigcup_{k \in \nat }\planset{k}$. 
The following definition introduces the notion
of history generated by a $k$-plan $\plan $ at an initial
state $s_0$. It is  the action-compatible $k$-history 
along which the agent executes the plan
$\plan $ starting at state $s_0$.
\begin{definition}[History generated  by a $k$-plan]
Let
$\effprecond=(\effprecondplus,\effprecondminus)$
be an action theory, $s_0 \in \stateset  $
and 
$\plan \in \planset{k}$.
Then, the history generated by plan  $\plan $
from state $s_0$
in conformity with  the action theory $\effprecond$ 
is the $k$-history $\history^{\pi,s_0,\effprecond }=
(\historyst^{\pi,s_0,\effprecond }, \historyact^{\pi,s_0,\effprecond })
$ 
such that:
\begin{align*}
& (i) \ \history^{\pi,s_0,\effprecond }\in \historyset{}(\effprecond ),\\
    & (ii) \ \historyst^{\pi,s_0,\effprecond }(0)=s_0,\\
    & (iii)\ \forall k' \text{ s.t. } 0 \leq  k' \leq k: \historyact^{\pi,s_0,\effprecond }(k')=\plan(k'),\\
\end{align*}

\end{definition}

Given 
a set of $\ltllogic$-formulas
$\Sigma  $,
we define  
$  \satset{\Sigma}{\plan }{s_0}{\effprecond}$
to be 
the set of formulas from $\Sigma $
that are guaranteed to be true  by 
the execution of 
plan $\plan $ at state $s_0$
under 
the action theory $\effprecond$.
That is, 
\begin{align*}
    \satset{\Sigma}{\plan }{s_0}{\effprecond}=
\big\{ \varphi \in \Sigma \suchthat
 \history^{\pi,s_0,\effprecond },
  0 \models \varphi
\big\}     .
\end{align*}

%As the following proposition indicates,
%if $\plan $
%is a $k$-plan, 
%then 
%the truth value of formulas does not change
% after time point $k+1 $
%along the history 
%$\history^{\pi,s_0,\effprecond }$.
%\begin{proposition}
% Let $\plan \in \planset{k}$
% and $\phi \in \langlogic_{\ltllogic}$. 
% Then, 
 %\begin{align*}
 % \history^{\pi,s_0,\effprecond },
 % k+1 \models \phi \text{ iff }
 % \forall k'\geq k+1 :
  %\history^{\pi,s_0,\effprecond },
  %k' \models \phi. 
 %\end{align*}
%\end{proposition}
%\begin{proof}
% ADD PROOF.
%\end{proof}

%Since a $k$-plan $\plan $ only changes the environment up to  time point $k+1 $, for every $\ltllogic$-formula $\phi$ we can find a $\wltllogic $-formula $\psi$ of polynomial size such that $\phi$ is satisfied on the history generated by $\plan $ if and only if $\psi$ is too. 
%Specifically,
%if
%$\plan \in \planset{k}$
 %with $k \in \nat $, 
%then
 %$\forall 
%\phi \in \langlogic_{\ltllogic},
%\exists \psi \in \langlogic_{\wltllogic}$
%such that the size of $\psi $
%is polynomial in the size of $\phi$ and: 
% \begin{align*}
%\history^{\pi,s_0,\effprecond },
 % 0 \models \phi \text{ iff }
%\history^{\pi,s_0,\effprecond },
%  0 \models \psi. 
% \end{align*}

\subsection{Moral Conflicts}

An ethical planning agent is likely to have multiple values that it wishes to satisfy when making plans. Some of these values will be ethical in nature (``do not harm humans''), and some may not be (``do not leave doors open''). However, the more values the robot has the more likely it is to  experience scenarios where it cannot satisfy all of its values with any given plan, and must violate some of them. In such a scenario, the agent must first work out which subsets of its value base are jointly satisfiable, and then which of those subsets it should choose to satisfy.

To this end we define a notion of a moral conflict (note that in line with \citet{LeviBook1990} we refer to any conflict between an agent's values as a ``moral conflict'' even if some or all of those values are not strictly moral/ethical in nature).

\begin{definition}[Moral Problem]\label{def:moralproblem}
A moral problem is a tuple $M = (\Omega, \gamma, s_0)$ where:
\begin{itemize}
    \item $\Omega \subseteq \langlogic_{\ltllogic}$ is a set of values (which may or may not be strictly moral in nature).
    \item $\effprecond=(\effprecondplus,\effprecondminus)$ is an action theory and $s_0$ is an initial state, as described above.
\end{itemize}
\end{definition}

\begin{definition}[Moral Conflict]\label{def:moralconflict}
A moral problem $M = (\Omega, \gamma, s_0)$ is a moral conflict if:
\begin{itemize}
    \item $\forall k \in \nat$, there is no $k$-plan $\pi$ such that $\satset{\Omega}{\plan }{s_0}{\effprecond} = \Omega$.
\end{itemize}
\end{definition}

In other words, a moral conflict occurs when it is not possible to satisfy all of our values with any given plan. In some cases, a moral conlict may not depend on any particular feature of the start state, but may result simply from the value set and action theory, or even the action theory alone. This allows us to define two further notions of moral problem.

\begin{definition}[Physical Moral Problem]\label{def:pmoralproblem}
A physical moral problem is a pair $(\Omega, \gamma)$ where:
\begin{itemize}
    \item $\Omega \subseteq \langlogic_{\ltllogic}$ is a set of values.
    \item $\effprecond$ is an action theory.
\end{itemize}
\end{definition}

\begin{definition}[Logical Moral Problem]\label{def:lmoralproblem}
A logical moral problem is a set of values $\Omega \subseteq \langlogic_{\ltllogic}$.
\end{definition}

We can also define moral conflict for these moral problems. A physical (logical) moral problem is a physical (logical) value conflict if for every possible start state $s_0$ (and every possible action theory $\gamma$), the resultant moral value problem $M = (\Omega, \gamma, s_0)$ is a moral conflict. By our definition, conflict mirrors the concept of necessity. Necessity would imply that \textit{every} possible plan satisfies all the values in $\Omega$, whereas conflict implies that \textit{no} plan satisfies all values. Thus it is interesting to note that our definitions of conflict have mirrors in philosophical literature \cite{sep-modality}. A physical moral conflict mirrors the notion of nomic necessity (necessary given the laws of nature) (at least from the perspective of the robot, for whom the action theory comprises the laws of nature) whereas a logical moral conflict mirrors the notion of logical necessity (necessary given the nature of logic).

If an agent is experiencing a moral conflict, one response would be to ``temporarily forget'' values until she has a satisfiable set.

\begin{definition}[Contraction]\label{def:contraction}
If $M = (\Omega, \gamma, s_0)$ is a moral problem and $M' = (\Omega', \gamma, s_0)$ is a moral problem, we say that $M'$ is a \textit{contraction} of $M$ if:
\begin{itemize}
    \item $\Omega' \subseteq \Omega$
    \item $M'$ is not a moral conflict.
\end{itemize}
\end{definition}

\begin{proposition}
If $M = (\Omega, \gamma, s_0)$ is a moral conflict, $\pi$ is a plan, and $\Omega' =  \satset{\Omega}{\plan }{s_0}{\effprecond}$ then $M' = (\Omega', \gamma, s_0)$ must be a valid contraction of $M$.
\end{proposition}

In this case, we refer to $M'$ as the contraction generated by $\pi$. This also illustrates that the current notion of contraction is unhelpful for an agent attempting to select a plan in a moral conflict, as all plans generate contractions. What would be helpful is some notion of a ``minimal'' or ``ideal'' contraction that sacrifices as few values as possible.

\begin{definition}[Minimal Contractions]\label{def:minimalcontractions}
If $M = (\Omega, \gamma, s_0)$ is a moral problem and $M' = (\Omega', \gamma, s_0)$ is a contraction of $M$, $M$ is:
\begin{itemize}
    \item A qual-minimal contraction if there is no contraction $M'' = (\Omega'', \gamma, s_0)$ such that $\Omega' \subset \Omega''$.
    \item A quant-minimal contraction if there is no contraction $M''$ such that $|\Omega'| < |\Omega''|$
\end{itemize}
\end{definition}

\begin{proposition}\label{prop:contractionsofproblems}
If $M = (\Omega, \gamma, s_0)$ is a moral problem and is not a moral conflict, then the only qual-minimal and quant-minimal contraction of $M$ is $M$.
\end{proposition}

For either notion of minimality, we will have cases where there are multiple minimal contractions of a given moral conflict. This can produce unintuitive results, as if there is some moral conflict with $\Omega = \{\text{``do not kill humans''},\\ \text{``do not leave the door open''}\}$ with contractions $\{\text{``do not kill humans''}\}$ and\\ $\{\text{``do not leave the door open''}\}$ then either notion of minimality will tell you that both contractions are ideal. On the other hand, it does seem that any stronger notion of minimality should at least respect qualitative minimality, since (intuitively), if plan $\pi_1$ fulfills all of the values fulfilled by $\pi_2$, and fulfills more values, then $\pi_1$ should be preferred to $\pi_2$.

\begin{proposition}
Given a moral conflict $M$, a contraction $M'$ is quant-minimal only if it is qual-minimal.
\end{proposition}

One way to resolve this is to recognise, in line with \citet{LeviBook1990}, that some of our values are only used as tiebreakers in cases of undecideability, and should not be considered directly alongside our more important values. In other words, our values exist in lexicographically ordered sets, where each set is examined only if the sets above cannot deliver a verdict.
 \subsection{Lexicographic Value Sets}
 
  Together with an action theory and an initial state, an agent's value base constitute an ethical planning domain.

\begin{definition}[Ethical planning domain]\label{def:plandomain}
An ethical planning domain
is a tuple $\pldomain=( \effprecond, s_0,\valueprof)$
where:
\begin{itemize}
    \item $\effprecond=(\effprecondplus,\effprecondminus)$
    is an action theory
    and $s_0$
    is an initial state,
    as specified above;
    \item
    $\valueprof = (\valueset{1}, \ldots,  \valueset{m} )$ is
    the agent's value base
   with 
$\valueset{k} \subseteq \langlogic_{\ltllogic}$
for every $1 \leq k \leq m$. 
\end{itemize}
\end{definition}
$\valueset{1}$
is the agent's set of values
with priority $1$, 
$\valueset{2}$
is the agent's set of values
with priority $2$, 
and so on. 
For notational convenience, 
given a value base 
$\valueprof= (\valueset{1}, \ldots,  \valueset{m} ) $,
we note $\mathit{dg}(\valueprof)$
its degree (or arity).

Agent's values 
are used to compute the  \emph{relative ideality} 
of plans,
namely, whether a plan $\plan_2$
is
at least as ideal as  another plan $\plan_1$.
Following \cite{DBLP:conf/atal/Lorini21},
we call 
\emph{evaluation} the operation of computing an ideality
ordering over plans 
from a value base.  
%We consider two lexicographic criteria of evaluation.
Building on classical preference representation languages \citet{Lang2004}, we define the following qualitative criterion of evaluation, noted $ \qualcomp{\pldomain }$, which compares two plans
lexicographically
on the basis of inclusion between sets of values. 
\begin{definition}[Qualitative ordering of plans]\label{def:ordering}
Let 
$\pldomain=( \effprecond, s_0,\valueprof)$ be
an ethical planning domain with
    $\valueprof = (\valueset{1}, \ldots,  \valueset{m} )$
    and  $ \plan_1, \plan_2\in \planset{}$.
    Then, $   \plan_1 \qualcomp{\pldomain } \plan_2$
    if and only if:
\begin{align*}
 & (i) \ \exists 1 \leq k \leq m
   \text{ s.t. }  \satset{\valueset{k}}{\plan_1 }{s_0}{\effprecond}\subset
           \satset{\valueset{k}}{\plan_2 }{s_0}{\effprecond},\\
         &  (ii) \ \forall 1 \leq k' < k,
           \satset{\valueset{k'}}{\plan_1 }{s_0}{\effprecond}=
           \satset{\valueset{k'}}{\plan_2 }{s_0}{\effprecond}.\\
         & or\\
         & (i) \ \forall 1 \leq k \leq m, \satset{\valueset{k}}{\plan_1 }{s_0}{\effprecond} =
           \satset{\valueset{k}}{\plan_2 }{s_0}{\effprecond}
\end{align*}
\end{definition}

Note that a quantitative criterion could also be defined by counting the number of satisfied values in each level and, in line with the previous definition, compare these values lexicographically.

The quantitative criterion,
noted $ \quantcomp{\pldomain }$, compares two plans 
lexicographically
on the basis of  comparative cardinality between sets of values.
\begin{definition}[Quantitative ordering of plans]\label{def:quantordering}
Let 
$\pldomain=( \effprecond, s_0,\valueprof)$ be
an ethical planning domain with
    $\valueprof = (\valueset{1}, \ldots,  \valueset{m} )$
    and      $ \plan_1, \plan_2\in \planset{}$.
    Then, $   \plan_1 \quantcomp{\pldomain } \plan_2$
    if and only if:
\begin{align*}
 & (i) \ \exists 1 \leq k \leq m
   \text{ s.t. }  |\satset{\valueset{k}}{\plan_1 }{s_0}{\effprecond}| <
           |\satset{\valueset{k}}{\plan_2 }{s_0}{\effprecond}|,\\
         &  (ii) \ \forall 1 \leq k' < k,
          | \satset{\valueset{k'}}{\plan_1 }{s_0}{\effprecond}|=
           |\satset{\valueset{k'}}{\plan_2 }{s_0}{\effprecond}|.\\
         & or\\
         & (i) \ \forall 1 \leq k \leq m, |\satset{\valueset{k}}{\plan_1 }{s_0}{\effprecond}| =
           |\satset{\valueset{k}}{\plan_2 }{s_0}{\effprecond}|
\end{align*}
\end{definition}

This allows us to define another notion of minimal contraction for a moral conflict, namely a minimal contraction with respect to a lexicographic value set. 

\begin{definition}[Lexicographic-minimal contraction]\label{def:lmincontraction}
If $M = (\Omega, \gamma, s_0)$ is a moral conflict, and $\valueprof = (\Omega_1, ... \Omega_m)$ is a value set such that $\cup \valueprof = \Omega$ then $M' = (\Omega', \gamma, s_0)$ is a $\valueprof$-qual-minimal contraction of $M$ if and only if:
\begin{align*}
 & (i) \ \Omega' \subseteq \Omega\\
 &  (ii) \ M \text{ is not a moral conflict}\\
 &  (iii) \ \text{If }M'' = (\Omega'', \gamma, s_0) \text{ is also a contraction of }M, \ \nexists k:\\
 & \ \ \ \ (a) \  1 \leq k \leq m \text{ and } \Omega' \cap \Omega_k \subset \Omega'' \cap \Omega_k\\
 & \ \ \ \ (b) \ \forall 1 \leq k' < k,\Omega' \cap \Omega_k' = \Omega'' \cap \Omega_k'.
\end{align*}
\end{definition}

Note that by combining definitions \ref{def:quantordering} and \ref{def:lmincontraction} we can define a notion of $\valueprof$-quant-minimal contraction.

\begin{proposition}\label{prop:respectsqminimal}
Given a moral conflict $M$, a contraction $M'$ is $\valueprof$-qual-minimal or $\valueprof$-quant-minimal only if it is qual-minimal.
\end{proposition}

 \subsection{Adding Desires}
 
 The behavior of
autonomous ethical agents
is driven not only by ethical
values aimed at promoting  the good for society 
but also
by their endogenous motivations, also  called
\emph{desires} or \emph{goals}. 
Following existing theories of ethical
preferences in philosophy, economics and logic \cite{Searle2001,HarsanyiEthics,DBLP:journals/flap/Lorini17},
we assume that (i) desires and values are competing 
motivational attitudes,
and (ii)
the agent's degree of morality is a function 
of its disposition to promote the fulfilment of
its values at the expense of the satisfaction
of its desires.  
The following definition
extends the notion 
of ethical planning domain by the notions of desire
and introduces the novel concept of degree of morality. 
\begin{definition}[Mixed-motive planning domain]\label{def:mixdomain}
A mixed-motive planning domain
is a tuple $\mixdomain=( \effprecond, s_0,\valueprof, \valueset{D},
\moraldegree{} )$
where 
\begin{itemize}
\item $( \effprecond, s_0,\valueprof)$
is an ethical
 planning domain (Definition \ref{def:plandomain});
 \item $\valueset{D} \subseteq \langlogic_{\ltllogic}$
 is the agent's set of desires or goals;
 \item $\moraldegree{} \in \{1, \ldots, \mathit{dg}(\valueprof )+1 \} $
 is the agent's degree of morality.
 \end{itemize}
\end{definition}

A mixed-motive planning domain induces
an ethical
planning domain whereby
the agent's set of desires
is treated as a set of values
whose priority level depends  
on the agent's degree of morality.
Specifically, the lower the agent's degree
of morality, the higher  the priority of 
the agent's set of desires
in the induced ethical planning domain. In many practical applications it is likely to be desirable to restrict the range of values that $\mu$ can take, in order to prevent (for example) the robot's goal from overriding its safety values.

\begin{definition}[Induced ethical planning domain]\label{def:inddomain}
Let  $\mixdomain=( \effprecond, s_0,\valueprof, \valueset{D},
\moraldegree{} )$ be a mixed-motive planning domain. 
The ethical planning domain 
induced by $\mixdomain$
is the tuple $\pldomain=( \effprecond, s_0,\valueprof')$
such that $\mathit{dg}(\valueprof' )=\mathit{dg}(\valueprof )+1$
with: 
\begin{align*}
    & (i) \ \valueset{\moraldegree{} }'= \valueset{D};\\
     & (ii)  \valueset{k}'= \valueset{k} \text{ for } 1 \leq k < \moraldegree{};\\
       & (iii) \valueset{k}'= \valueset{k-1} \text{ for } \moraldegree{} <k \leq \mathit{dg}(\valueprof )+1.
\end{align*}
\end{definition}

\section{An Example}

Consider a blood delivery robot in a hospital. The robot mostly makes deliveries between different storage areas, and sometimes delivers blood to surgeries. The robot may have to deal with various kinds of obstacles to complete its deliveries, but we will consider only one: people blocking the robot. The robot has two methods to resolve this obstacle, it can ask for them to move and then wait for them to move ($\rask$), or it can use a loud air-horn to ``force'' them to move ($\horn$). Once the person has moved, the robot can reach its destination ($\move$).
We suppose that the robot can tell some things about its environment, it knows if it is blocked ($\blocked$), if it is near the operating theatre ($\surgery$) and if it has reached its destination ($\destination$). We can then define the action model as follows:
%$\gamma^+$ and $\gamma^-$ as follows:

%%%DANGER: check if it makes problems
\vspace{-0.3cm}

\begin{align*}
\gamma^+(\move,\destination) &= \neg \blocked\\
\gamma^-(\rask,\blocked) &= \blocked\\
\gamma^+(\rask,\wait) &= \top\\
\gamma^-(\horn,\blocked) &= \blocked\\
\gamma^+(\horn,\annoyed) &= \top\\
\gamma^+(\horn,\dangerous) &= \surgery\\
\text{otherwise,} \gamma^\pm(a,p) &= \bot
\end{align*}

%\vspace{-0.2cm}

The propositions $\wait$, $\annoyed$ and $\dangerous$ are used to keep track of the robot's actions, we suppose that using the horn near the operating theatre is dangerous.
The values and desires of the robot can be presented as follows:

%%%DANGER: check if it makes problems
\vspace{-0.3cm}

\begin{align*}
\overline{\Omega} &= \{\Omega_1,\Omega_2\}\\
\Omega_1 &= \{\henceforth \neg \dangerous\}\\
\Omega_2 &= \{\henceforth \neg \annoyed\}\\
\Omega_D &= \{\eventually \destination, \eventually (\destination \land \neg \wait)\}\\
\end{align*}

%%%DANGER: check if it makes problems
\vspace{-0.2cm}

In words, the robot's goal is to reach its destination without delays, with primary value to never do anything dangerous, and secondary value to neverbe annoying. Let $\valueprof'$ be the value set induced by $\valueprof$, $\Omega_D$ and $\mu = 3$.

Now we can compare the following 2-plans $\pi_1 = (\rask, \move)$ and $\pi_2 = (\horn, \move)$. If we assume that in the initial state the robot is blocked but far from an operating theatre, we can represent the histories generated from these plans as follows (each block contains exactly the propositions that are true in that state):

\medskip

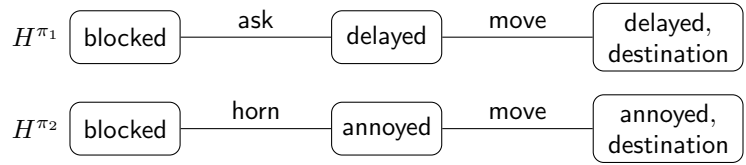
\tikzstyle{block} = [rectangle, draw, text width=3.5em, text centered, rounded corners, minimum height=2em]
\tikzstyle{block2} = [rectangle, draw, text width=5em, text centered, rounded corners, minimum height=2em]
\tikzstyle{line} = [draw]
\tikzstyle{line2} = [draw, dashed]

\begin{tikzpicture}[node distance=1cm, auto]
    \node (init) {};
    \node [block, label=left:{$\history^{\pi_2}$}] (A) {$\blocked$};
    \node [block, right=2cm of A] (B) {$\annoyed$};
    \node [block2, right=2cm of B] (C) {$\annoyed$, $\destination$};
    \node [block, label=left:{$\history^{\pi_1}$}, above=0.5cm of A] (D) {$\blocked$};
    \node [block, right=2cm of D] (E) {$\wait$};
    \node [block2, right=2cm of E] (F) {$\wait$, $\destination$};
    \path [line] (A) -- node [text width=2.5cm,midway,above,align=center ] {$\horn$} (B);
    \path [line] (B) -- node [text width=2.5cm,midway,above,align=center ] {$\move$} (C);
    \path [line] (D) -- node [text width=2.5cm,midway,above,align=center ] {$\rask$} (E);
    \path [line] (E) -- node [text width=2.5cm,midway,above,align=center ] {$\move$} (F);
\end{tikzpicture}

\medskip

In this case $\satset{\valueprof'}{\plan_1 }{s_0}{\effprecond} = \{\henceforth \neg \dangerous, \henceforth \neg \annoyed, \eventually \destination\} = \Omega^1$ whereas $\satset{\valueprof'}{\plan_2 }{s_0}{\effprecond} = \{\henceforth \neg \dangerous, \eventually \destination, \eventually(\destination \land \neg \wait)\} = \Omega^2$.
Therefore $\pi_1$ will be preferred to $\pi_2$. However, if we change the morality level to $2$, perhaps to represent an urgent delivery to an ongoing surgery, then we see that the robot will choose plan $\pi_2$ rather than $\pi_1$. This illustrates how we can adjust the morality level of the robot to reflect the urgency of its goals. 
If we move the example to the operating theatre (so now $\surgery \in s_0$ instead of $\neg \surgery \in s_0$), then the robot would not sound its horn even if the delivery was urgent, as $\Omega_1$ still overrides $\Omega_D$. 
This also means that for this robot we should restrict $\mu$ to $\{2,3\}$ to ensure that values in $\Omega_1$ are always prioritised over goals. Furthermore, notice that for any lexicographic value structure containing exactly these values and goals, the set of non-dominated plan will always be either $\{\pi_1\}$, $\{\pi_2\}$ or $\{\pi_1, \pi_2\}$ since $\Omega^1$ and $\Omega^2$ are exactly the qual-minimal contractions of $\cup \valueprof'$.

\section{Computational Complexity}

%\subsection{Comparison Problem}
In this section we initiate the study of the computational complexity of ethical planning in our setting. We borrow our terminology from the work of \citet{Lang2004} on compact preference representation, but the problems we study have obvious counterparts in the planning literature, as should be clear from the proofs. In the interest of space all proofs can be found in the appendix.

We begin by studying the problem \textsc{Conflict}, which determines if a moral problem is also a moral conflict.

	\begin{quote}
		\noindent \textsc{Conflict}\\
		\hspace*{-1em} \indent\textit{Input:} Moral problem $M = (\Omega, \gamma, s_0)$\\
		\hspace*{-1em}\textit{Question:} Is there some $k \in \nat$ such that there is a $k$-plan $\pi'$ such that $\satset{\Omega}{\pi }{s_0}{\effprecond} = \Omega$?
	\end{quote}

\begin{theorem}\label{thm:conflict}
\textsc{Conflict} is PSPACE-complete.
\end{theorem}	

We then study the case of contractions, in particular, determining if a given moral problem is a qual-minimal contraction.

	\begin{quote}
		\noindent \textsc{Minimal-Contraction}\\
		\hspace*{-1em} \indent\textit{Input:} Moral problem $M = (\Omega, \gamma, s_0)$, moral problem $M' = (\Omega', \gamma, s_0)$\\
		\hspace*{-1em}\textit{Question:} Is $M'$ a qual-minimal contraction of $M$?
	\end{quote}

\begin{theorem}\label{thm:contractions}
\textsc{Minimal-Contraction} is PSPACE-complete.
\end{theorem}	

Neither of these results are particularly technically advanced, indeed \textsc{Conflict} is almost exactly equivalent to PLANSAT from classical planning \cite{STRIPS}. The purpose of these results is to indicate that quite apart from the issue of how a robot should select the best option when faced with a moral conflict, the task of identifying that the robot is facing a moral conflict and determining all of its options is extremely computationally difficult. On the subject of planning, we begin by studying the problem \textsc{Comparison}, which takes as input an initial state $s_0$, an ethical planning domain ${\Delta}$, two $k$-plans $\pi_1$ and $\pi_2$, and asks whether  $\pi_1 \qualcomp{\pldomain } \pi_2$. 
Despite the apparent complexity of our setting this problem can be solved efficiently:

\begin{theorem}\label{thm:comparison}
\textsc{Comparison} is in \textsc{P}.
\end{theorem}

We then move to the problem of non-dominance, i.e., the problem of determining if given a $k$-plan $\pi_1$ there exists a better $k$-plan wrt. $\qualcomp{\pldomain }$. 

	\begin{quote}
		\noindent \textsc{Non-dominance}\\
		\hspace*{-1em} \indent\textit{Input:} Ethical planning domain $\Delta = (\gamma, s_0, \valueprof)$, $k\in \mathbb N$, $g$-plan $\pi$ for $g\leq k$\\
		\hspace*{-1em}\textit{Question:} is there a $k$-plan $\pi'$ such that $\pi \qualcomp{\pldomain } \pi'$ and $\pi' \not\qualcomp{\pldomain } \pi$?
	\end{quote}
	
We show that this problem, as most instances of classical planning satisfaction, is PSPACE-complete:

\begin{theorem}\label{thm:nondominance}
\textsc{Non-Dominance} is PSPACE-complete.
\end{theorem}

\begin{proposition}
Given an ethical planning domain $\Delta = (\gamma, s_0, \valueprof)$, a $k$-plan $\pi$ and $S = \satset{\cup \valueprof}{\plan }{s_0}{\effprecond}$ $\pi$ is non-dominated for $\Delta$ if and only if $M = (S, \gamma, s_0)$ is a $\valueprof$-qual-minimal contraction for $(\cup \valueprof, \gamma, s_0)$.
\end{proposition}
Theorems~\ref{thm:comparison} and~\ref{thm:nondominance} are to be interpreted as baseline results showing the computational feasibility of our setting for ethical planning with $\ltllogic$. One clear direction for future work would expand on the computational complexity analysis, identifying tractable fragments and exploring their expressivity in ethical applications.

%\subsection{Why did you do that?}
%commented out since we have rather short sections

An important property for an ethical planner is \emph{explainability}. While explaining why a particular plan was chosen is difficult to do succinctly (even for humans),
a simpler problem is to explain why the chosen plan was better than another proposed alternative.
Our approach enables this in a way that is both computationally straightforward and intuitively understandable to humans, since by the lexicographic ordering of plans there always exists a single value or set of values that decides between two plans.

%\noteug{If we need space we can stop here.}
%For example, if a technician wants to know why the robot picked plan $\{\rask, \move\}$ instead of plan $\{\horn, \move\}$, it should be relatively straightforward for the robot to give the answer ``because the second plan would have caused an annoyance''. 
%This is likely to be a more appealing and easily understood explanation than (for instance) an ethical utility score for each plan, particularly to a lay audience. 
%How appealing these explanations are will in part depend on getting the ``correct'' order of values in $\valueprof$.
%\noteug{We need references here on explanation in planning, especially those that make use of utilities.}

\section{Conclusions}
We put forward a novel setting for ethical planning obtained by combining a simple logical temporal language with lexicographic preference modelling.
Our setting applies to planning situations with a single agent who has deterministic and instantaneous actions to be performed sequentially in a static and known environment.  
Aside from the addition of values, our framework differs from classical planning in two aspects, by having multiple goals and by allowing temporal goals. 
%In terms of the scenarios that this system could be applied to, 
In particular, the expressiveness of LTL means that we can express a wide variety of goals and values, including complex temporal values such as ``if the weather is cold, close external doors immediately after opening them'', with a computational complexity equivalent to that of standard planners.
As a limitation, the system is less able to express values that tend to be satisfied by degree rather than absolutely or not at all. 
%For example, we might think it matters \textit{by how much} a car is breaking the speed limit rather the mere fact that it \textit{is} speeding. There may also be cases where some values cannot be clearly ordered (better to be a bit late than very annoying, but better to be a bit annoying than very late). Both of these issues can be dealt with by splitting values (so we have the values ``don't be mildly annoying'' and ``don't be very annoying''), but this risks an explosion in the number of values.
%The system that we have presented here could be used in various ways to aid automated ethical planning. Our system cannot be straightforwardly applied to most planning algorithms as it does not have a single goal, and there are no values that are guaranteed to be satisfied by a non-dominated plan.
Among the multiple directions for future work that our definitions open, we plan to study the multi-agent extension with possibly conflicting values among agents, moving from plans to strategies (functions from states or histories to actions), from complete to incomplete information, and, most importantly, test our model by implementing it in simple robotics scenarios. Furthermore, given the computational complexity of \textsc{Conflict}, \textsc{Mininal-Contraction} and \textsc{Non-Dominance}, it may often be the case that in practical applications we cannot guarantee finding a non-dominated plan. Therefore, it would be valuable to find more tractable algorithms that at least guarantee some degree of approximation of a non-dominated plan, or restrictions (likely to the language or action theory) that improve tractability of the problem.
%of plan verification and finding.

%\noteug{Tim: I am not sure I understand the idea of comparing sets of plans in the paragraph that follows. Since comparison is polynomial if the set of plans is not too big I can easily find the best one. As a general comment, we should probably list here a number of high-level future work that showcase the generality of our model, rather than going into the details of a single problem.}
%One potential use for our system is to compare sets of plans. For example, we could use a separate aglorithm to generate plans for $\Omega_D$ and then compare those plans (along with $\pi_{skip}$) to see which is ethically preferable. Another option would be something like a binary search. This would mean guessing a combination of values that are satisfiable and treating them as a single conjunctive goal. This would enable the use of standard LTL planners, and then we set this combination of values as either an upper or lower bound depending on whether or not the conjunction is reachable. Furthermore, in many cases we could determine upper and lower bounds in advance. For example, in the blood-delivery example above, we know that $\{\destination, \neg \wait, \neg \annoyed\}$ is unreachable if $\blocked \in s_0$. This approach will work best when using small or singleton $\Omega_i$ and/or $\quantcomp{\pldomain }$ rather than $\qualcomp{\pldomain }$.

\bibliography{biblio.bib}

\begin{thebibliography}{10}
\providecommand{\url}[1]{\texttt{#1}}
\providecommand{\urlprefix}{URL }
\providecommand{\doi}[1]{https://doi.org/#1}

\bibitem{AliliEtAl2008}
Alili, S., Alami, R., Montreuil, V.: A task planner for an autonomous social
  robot. In: Proceedings of the 9th International Symposium on Distributed
  Autonomous Robotic Systems (DARS). Springer (2008)

\bibitem{GenEth}
Anderson, M., Anderson, S.L.: Geneth: a general ethical dilemma analyzer.
  Paladyn (Warsaw)  \textbf{9}(1),  337--357 (2018)

\bibitem{DBLP:journals/pieee/ArkinUW12}
Arkin, R.C., Ulam, P., Wagner, A.R.: Moral decision making in autonomous
  systems: Enforcement, moral emotions, dignity, trust, and deception.
  Proceedings of the {IEEE}  \textbf{100}(3),  571--589 (2012)

\bibitem{AwadEtAlarXiv2022}
Awad, E., Levine, S., Loreggia, A., Mattei, N., Rahwan, I., Rossi, F.,
  Talamadupula, K., Tenenbaum, J.B., Kleiman{-}Weiner, M.: When is it
  acceptable to break the rules? {K}nowledge representation of moral judgement
  based on empirical data. CoRR  \textbf{abs/2201.07763} (2022),
  \url{https://arxiv.org/abs/2201.07763}

\bibitem{DBLP:journals/ai/BenzmullerPT20}
Benzm{\"{u}}ller, C., Parent, X., van~der Torre, L.W.N.: Designing normative
  theories for ethical and legal reasoning: Logi{KE}y framework, methodology,
  and tool support. Artificial Intelligence  \textbf{287},  103--348 (2020)

\bibitem{DBLP:conf/atal/BerrebyBG17}
Berreby, F., Bourgne, G., Ganascia, J.: A declarative modular framework for
  representing and applying ethical principles. In: Proceedings of the 16th
  Conference on Autonomous Agents and MultiAgent Systems (AAMAS) (2017)

\bibitem{STRIPS}
Bylander, T.: The computational complexity of propositional {STRIPS} planning.
  Artif. Intell.  \textbf{69}(1-2),  165--204 (1994).
  \doi{10.1016/0004-3702(94)90081-7},
  \url{https://doi.org/10.1016/0004-3702(94)90081-7}

\bibitem{HandbookEthics2007}
Copp, D.: The Oxford Handbook of Ethical Theory. Oxford University Press (2007)

\bibitem{DennisFSW16}
Dennis, L.A., Fisher, M., Slavkovik, M., Webster, M.: Formal verification of
  ethical choices in autonomous systems. Robotics and Autonomous Systems
  \textbf{77},  1--14 (2016)

\bibitem{Dennis}
Dennis, L.A., del Olmo, C.P.: A defeasible logic implementation of ethical
  reasoning. In: First International Workshop on Computational Machine Ethics
  (CME-2021) (2021)

\bibitem{EvansK}
Evans, K., de~Moura, N., Chauvier, S., Chatila, R., Dogan, E.: Ethical decision
  making in autonomous vehicles: The av ethics project. Science and engineering
  ethics  \textbf{26}(6),  3285--3312 (2020)

\bibitem{GiacomoV13}
Giacomo, G.D., Vardi, M.Y.: Linear temporal logic and linear dynamic logic on
  finite traces. In: Rossi, F. (ed.) {IJCAI} 2013, Proceedings of the 23rd
  International Joint Conference on Artificial Intelligence, Beijing, China,
  August 3-9, 2013. pp. 854--860. {IJCAI/AAAI} (2013),
  \url{http://www.aaai.org/ocs/index.php/IJCAI/IJCAI13/paper/view/6997}

\bibitem{HarsanyiEthics}
Harsanyi, J.: Utilitarianism and beyond. In: Sen, A.K., Williams, B. (eds.)
  Morality and the theory of rational behaviour. Cambridge University Press,
  Cambridge (1982)

\bibitem{Jenkin}
Jenkins, R., Talbot, B., Purves, D.: When robots should do the wrong thing. In:
  Robot Ethics 2.0. Oxford University Press, New York (2017)

\bibitem{sep-modality}
Kment, B.: {Varieties of Modality}. In: Zalta, E.N. (ed.) The {Stanford}
  Encyclopedia of Philosophy. Metaphysics Research Lab, Stanford University,
  {S}pring 2021 edn. (2021)

\bibitem{Lang2004}
Lang, J.: Logical preference representation and combinatorial vote. Annals of
  Mathematics and Artificial Intelligence  \textbf{42}(1-3),  37--71 (2004)

\bibitem{LeviBook1990}
Levi, I.: Hard Choices: Decision Making Under Unresolved Conflict. Cambridge
  University Press (1990)

\bibitem{LoreggiaEtAlAAMAS2018}
Loreggia, A., Mattei, N., Rossi, F., Venable, K.B.: On the distance between
  cp-nets. In: Proceedings of the 17th International Conference on Autonomous
  Agents and MultiAgent Systems (AAMAS) (2018)

\bibitem{LoreggiaEtAl2017}
Loreggia, A., Rossi, F., Venable, K.B.: Modelling ethical theories compactly.
  In: The Workshops of the The Thirty-First {AAAI} Conference on Artificial
  Intelligence (2017)

\bibitem{LoriniMoral}
Lorini, E.: A logic for reasoning about moral agents. Logique \& Analyse
  \textbf{58}(230),  177--218 (2015)

\bibitem{DBLP:journals/flap/Lorini17}
Lorini, E.: Logics for games, emotions and institutions. {FLAP}  \textbf{4}(9),
   3075--3113 (2017)

\bibitem{DBLP:conf/atal/Lorini21}
Lorini, E.: A logic of evaluation. In: Proceedings of the 20th International
  Conference on Autonomous Agents and Multiagent Systems (AAMAS). pp. 827--835.
  ACM (2021)

\bibitem{sep-ethics-ai}
Müller, V.C.: {Ethics of Artificial Intelligence and Robotics}. In: Zalta,
  E.N. (ed.) The {Stanford} Encyclopedia of Philosophy. Metaphysics Research
  Lab, Stanford University, {S}ummer 2021 edn. (2021)

\bibitem{Pnueli1977}
Pnueli, A.: The temporal logic of programs. In: Proceedings of the 18th Annual
  Symposium on Foundations of Computer Science (FOCS) (1977)

\bibitem{Powers}
Powers, T.M.: Deontological machine ethics. In: Anderson, M., Anderson, S.L.,
  Armen, C. (eds.) Association for the Advancement of Artificial Intelligence
  Fall Symposium Technical Report (2005)

\bibitem{RossiMatteiAAAI2019}
Rossi, F., Mattei, N.: Building ethically bounded {AI}. In: The Thirty-Third
  {AAAI} Conference on Artificial Intelligence (AAAI) (2019)

\bibitem{Searle2001}
Searle, J.: Rationality in Action. Cambridge University Press, MIT Press (2001)

\bibitem{SenEthicsBook}
Sen, A.: On Ethics and Economics. Basil Blackwell (1987)

\bibitem{Vallor}
Vallor, S.: Technology and the Virtues: A Philosophical Guide to a Future Worth
  Wanting. Oxford University Press, New York (2016)

\bibitem{DBLP:journals/cogsr/VanderelstW18}
Vanderelst, D., Winfield, A.F.T.: An architecture for ethical robots inspired
  by the simulation theory of cognition. Cognitive Systems Research
  \textbf{48},  56--66 (2018)

\bibitem{YuEtAlIJCAI2018}
Yu, H., Shen, Z., Miao, C., Leung, C., Lesser, V.R., Yang, Q.: Building ethics
  into artificial intelligence. In: Proceedings of the 27th International Joint
  Conference on Artificial Intelligence (IJCAI) (2018)

\end{thebibliography}

\newpage

\section*{Appendix: Missing proofs}
\textbf{Proof of Proposition~\ref{prop:respectsqminimal}}

\begin{proof}
Let $M = (\Omega, \gamma, s_0)$ and let $\valueprof$ be a lexicographic ordering of $\Omega$ of degree $m$. Suppose $M' = (\Omega, \gamma, s_0)$ is a $\valueprof$-qual-minimal contraction of $M$. Suppose for contradiction that $M'$ is not qual-minimal. Then there exists some contraction $M'' = (\Omega'', \gamma, s_0)$ such that $\Omega' \subset \Omega''$. Therefore there exists some value $\phi \in \Omega$ such that $\phi \in \Omega''$ and $\phi \notin \Omega'$. Let $p$ be the priority level of $\phi$ in $\valueprof$ (so $\phi \in \Omega_p$).

Since $\Omega' \subset \Omega''$, we know that for all $1 \leq k \leq m$, either $\Omega' \cap \Omega_k \subset \Omega'' \cap \Omega_k$ or $\Omega' \cap \Omega_k = \Omega'' \cap \Omega_k$. We also know that $\Omega' \cap \Omega_p \subset \Omega'' \cap \Omega_p$. Therefore there must be some $p' \leq p$ such that:$1 \leq p' \leq m \text{ and } \Omega' \cap \Omega_{p'} \subset \Omega'' \cap \Omega_{p'}$ and $\forall 1 \leq k' < p',\Omega' \cap \Omega_k' = \Omega'' \cap \Omega_k'$.

By a very similar method, we can derive a contradiction if we suppose that $M'$ is $\valueprof$-quant-minimal but not qual-minimal.
\end{proof}

\textbf{Proof of Theorem~\ref{thm:conflict}}

\begin{proof}
To show that \textsc{Conflict} is PSPACE-hard, we show a reduction from the classical planning problem PLANSAT for propositional STRIPS planning \cite{STRIPS}. In this problem we have a set of \textit{conditions} (propositions) that can be true of false, an initial state which is a collection of conditions, a set of \textit{operators} (actions) that have preconditions and postconditions as sets of satisfiable conjunctions of positive and negative conditions, and a single \textit{goal} which is a conjunction of positive and negative conditions. We then attempt to find a finite sequence of operators that acheives the goal from the starting state. For a more complete description and complexity results, see \cite{STRIPS}.

To perform the reduction, set $\Omega = \{\omega\}$ where $\omega$ is our goal. Creating an action theory $\gamma$ from the set of operators and a start state $s_0$ can be done in polynomial time. Then \textsc{Conflict} applied to $(\Omega, \gamma, s_0)$ returns \textsc{true} if and only if PLANSAT would return \textsc{false}.

To show that \textsc{Conflict} is PSPACE-complete, we show a reduction from \textsc{Conflict} to PLANSAT. Given a moral problem $M = (\Omega, \gamma, s_0)$ define $\phi$ as the conjunction of all formulas in $\Omega$, then set $\phi$ as our goal. We can generate a set of operators from $\gamma$ and an initial state from $s_0$ in polynomial time. Then PLANSAT returns \textsc{true} if and only if \textsc{Conflict} would return \textsc{false}.

\end{proof}

\par \noindent \textbf{Proof of Theorem~\ref{thm:contractions}}

\begin{proof}
To show that \textsc{Minimal-Contraction} is PSPACE-hard, we show a reduction from \textsc{Conflict }. Given a value problem $N$, set $M = M' = N$. Then by proposition \ref{prop:contractionsofproblems}, \textsc{Minimal-Contraction} will return \textsc{true} if and only if \textsc{Conflict} would return \textsc{false}.

To show that \textsc{Minimal-Contraction} is PSPACE-complete, we provide a basic algorithm that uses polynomial space. First, use \textsc{Conflict} to check if $M'$ is a conflict, if it is, return \textsc{false}. If not, let $A = \Omega \backslash \Omega'$. For each $a \in A$, run \textsc{Conflict} on $(\Omega' \cup {a}, \gamma, s_0)$. If \textsc{Conflict} returns \textsc{false} on any of these checks, return \textsc{false}, else, return \textsc{true}.
\end{proof}

\par \noindent \textbf{Proof of Theorem~\ref{thm:comparison}}
\begin{proof}
Recall that to compare two plans $\pi_1$ and $\pi_2$ we need an ethical planning domain $\Delta = (\gamma, s_0, \valueprof)$ where $\gamma$ is an action theory, $s_0$ is an initial state and $\valueprof$ is a value base. 
Following Definition~\ref{def:ordering}, plan $\pi_1$ is better than $\pi_2$ if the history generated by $\pi_1$ is lexicographically preferred to the history generated by $\pi_2$ according to the ranked values in $\valueprof$.

We begin by showing that generating the unique history associated to a $k$-plan can be done in polynomial time. Then we show that evaluating an $\ltllogic$ formula over this history can be done in polynomial time, concluding that \textsc{Comparison} is in P since we can give an answer by model checking all formulas in $\valueprof$ over the histories generated by the two plans.

Generating $\historyact$ associated to $\pi$ can be done by making a copy of the plan $\pi$ that returns $\nullact$ whenever the input is greater than $k$. This can be done in polynomial time.
We then set $\historyst(0) = s_0$, then for each $\historyst(i)$ we use $\gamma$ to generate $\historyst(i+1)$ in polynomial time by model checking all formulas in $\gamma$. We only have to do this $k$ times.

Let us now show that $\ltllogic$ formulas can be checked in polynomial time on the history generated by a $k$-plan. 
Suppose we have an $\ltllogic$ formula $\phi$ of length $n$, and a history $\history$ associated with some $k$-plan $\pi$. We proceed by strong induction on $n$. For the purpose of this proof, suppose an algorithm that takes $\phi$ and $\history$ as inputs. 
Let $\history_j$ be the history such that $\historyactj(i) = \historyact(i+j)$ and $\historystj(i) = \historyst(i+j)$.

\textbf{Base case.} Suppose n = 1, then $\phi = p$ for some $p \in \propset$. Then we can determine if $p \in \historyst(0)$ in polynomial time.

\textbf{Inductive step.} Suppose $n > 1$ and that the claim holds for all $m < n$. Then we have several options for $\phi$.

\begin{enumerate}
    \item $\phi = \neg \psi$. Then by inductive hypothesis we can determine in polynomial time if $H \vDash \psi$ and thus if $H \vDash \phi$.
    \item $\phi = \psi \land \chi$. Then we can determine in time $P$ if $\history \vDash \psi$ and $\history \vDash \chi$.
    \item $\phi = \nexttime \psi$. Then we can determine (in polynomial time) if $\history_1 \vDash \psi$.
    \item $\phi = \until{\psi}{\chi}$. Then for $0 < j < k$ we can determine if $\history_0, \history_1, ... , \history_{j-1} \vDash \psi$ and $\history_j \vDash \chi$ in polynomial time. Therefore this whole process can be done in polynomial time.
    
\end{enumerate}

To conclude, suppose we have an ethical planning domain $\Delta = (\gamma, s_0, \valueprof)$ and $k$-plans $\pi_1$ and $\pi_2$. 
By the previous steps we can generate $\history^{\pi_1}$ and $\history^{\pi_2}$, and for each $\omega \in \valueprof$ we can determine whether $\history^{\pi_1}$ and $\history^{\pi_2} \vDash \omega$ in polynomial time. Therefore evaluating for every possible value can be done in polynomial time.
Determining $\qualcomp{\pldomain}$ 
%or $\quantcomp{\pldomain}$ 
involves checking every value a maximum of once, so we can conclude that $\textsc{Comparison}$ is in $P$.
\end{proof}

\textbf{Proof of Theorem~\ref{thm:nondominance}}

\begin{proof}
To show that \textsc{Non-Dominance} is PSPACE-hard, we show a reduction from the classical planning problem $\textsc{PLANMIN}$ for propositional STRIPS planning \cite{STRIPS}. In this problem we have a set of \textit{conditions} (propositions) that can be true of false, a set of \textit{operators} (actions) that have preconditions and postconditions as sets of satisfiable conjunctions of positive and negative conditions, and a single \textit{goal} which is a satisfiable conjunction of positive and negative conditions. We then attempt to find a sequence of $k$ or less operators that acheives the goal from the starting state. For a more complete description and complexity results, see \cite{STRIPS}.

To perform the reduction, set $\valueprof = \Omega_1$ where $\Omega_1 = \{\omega\}$ where $\omega$ is our goal. Creating an action theory from the set of operators can be done in polynomial time. Then, generate a random $k$-plan $\pi$ and check if $\history^\pi \vDash \omega$, if it does then we are done. If it does not then non-dominance applied to $\pi$ is equivalent to PLANMIN.

To show that \textsc{Non-Dominance} is PSPACE-complete we provide a basic algorithm that uses polynomial space. Given $\Delta = (\gamma, s_0, \valueprof)$ and $k$-plan $\pi$, check every possible plan $\pi'$ for $\gamma$ and $s_0$ and check \textsc{Comparison} against $\pi$. 
Terminate once a plan is found that dominates $\pi$ or once all plans have been checked.
This algorithm only needs two plans in memory at any one time ($\pi$ and the plan being compared to $\pi$), and therefore it only requires polynomial space.
\end{proof}

\end{document}